\documentclass[9pt,twocolumn]{opticajnl}
\journal{opticajournal}

\setboolean{shortarticle}{false}   
\setboolean{displaycopyright}{false}

\usepackage{booktabs}
\usepackage{graphicx}
\usepackage{amsmath}
\usepackage{array}
\usepackage{hyperref}
\usepackage{needspace}

\usepackage{iftex}
\ifPDFTeX
  \newcommand{\tablefont}{\fontfamily{pag}\selectfont}
\else
  \usepackage{fontspec}
  \newfontfamily\gifont{Glacial Indifference}[
    Path=fonts/,
    Extension=.ttf,
    UprightFont=GlacialIndifference-Regular,
    BoldFont=GlacialIndifference-Bold,
    Ligatures=TeX]   
  \newcommand{\tablefont}{\gifont}
\fi

\dates{This paper was peer-reviewed and presented at NetMob 2026 (Data Challenge Track).}

\title{Detecting bus delays and service anomalies in Niter\'oi with autoencoder
latent feature representations}

\author{Matheus Germano}
\author{Claudio E. C. Campelo}

\affil{Systems and Computing Department, Federal University of Campina Grande (UFCG), Campina Grande, PB, Brazil}

\AtBeginDocument{\renewcommand{\nocitefullrefs}[1]{}}

\begin{document}

\twocolumn[\begin{@twocolumnfalse}
\maketitle
\vspace{-6pt}
\rule{\textwidth}{0.4pt}\vspace{4pt}
{\small\noindent\textbf{Abstract.} Bus riders judge service by two failures: single trips that run late, and days
on which the whole system misbehaves. We ask whether one unsupervised
representation of raw GPS telemetry can serve both diagnoses on the NetMob 2026
Niter\'oi bus dataset. An autoencoder compresses each trip into a 64-dimensional
latent vector; the frozen encoder feeds supervised delay classifiers, and the
reconstruction error serves as an anomaly score. Two findings are negative.
First, the latent representation does not improve delay detection: gradient
boosting on per-trip summary features reaches $0.846 \pm 0.013$ ROC-AUC against
$0.838 \pm 0.016$ for the best latent setup, and the gap becomes significant
once the preprocessing filters are relaxed. Second, two of the three external
checks of the anomaly score were circular. Retraining without rainfall removes
the rain association ($5.4\%$ against $5.0\%$ flagged, $p = 0.84$) and an
apparent citywide disruption on 23 March. What survives is a day-level signal
stable across three autoencoders (Spearman $\rho \geq 0.96$), driven mostly by
day type, with degraded-telemetry days still separable within weekday and
weekend strata. The alarm threshold limits detection more than the score does:
a pooled percentile cut flags $5.7\%$ of trips on the worst degraded day, while
re-scaling within each bus line flags $29.3\%$. The delay model rests almost
entirely on one feature, no latent dimension is individually interpretable,
and 22 chronic line-directions carry half of all delayed trips.}
\vspace{4pt}\rule{\textwidth}{0.4pt}
\vspace{10pt}
\end{@twocolumnfalse}]

\section{Introduction}
\label{sec:intro}

Riders judge a bus system by two failures: trips that run late, and days when
service as a whole misbehaves; travellers value reliability on a par with travel
time itself~\cite{Bates2001reliability}. We ask whether a single unsupervised
representation of raw GPS telemetry can support both diagnoses at once on the
NetMob 2026 challenge data from Niter\'oi~\cite{netmob26}. Our approach
transfers the latent feature representation learning pipeline of Duncan and
Chen~\cite{Duncan2023}, originally built to detect internet censorship in
network traffic: an autoencoder (AE) is trained to compress each trip into a
low-dimensional latent vector, the frozen encoder then feeds lightweight
supervised classifiers for \emph{delay detection}, and the same AE's
reconstruction error doubles as an unsupervised \emph{service anomaly} score.
Autoencoders are long established~\cite{Hinton2006autoencoder} and have been
applied to anomalous bus trajectories~\cite{Zhang2021bus}; here we reuse one
label-free representation for both tasks. This is particularly useful when
timetable-based comparisons are unreliable --- for example, when schedule joins
are missing or outdated, when service runs at variable intervals, or when trips
are evaluated before they are complete.

\subsection{Motivation for a learned model}
\label{sec:why}

The General Transit Feed Specification (GTFS)~\cite{gtfs} is the standard open
format in which transit agencies publish their timetables. A feed is a set of
CSV files describing the routes, the stops, the trips run on each route, and the
scheduled arrival time at every stop of every trip. Where such a feed exists, a
deterministic comparison against it would identify delays directly, so a learned
classifier needs justifying. 

GTFS is used in this work to construct the delay labels and is excluded from the model inputs.
We keep the two apart on purpose, to block leakage. A model handed
schedule deviation as a feature would simply recover a label computed from that
same input, and would then be measuring arithmetic instead of the quality of the
representation. Every input channel is derived from GPS telemetry,
weather, or headway. None is derived from the schedule.

Given that separation, the learned model targets three operating conditions a
deterministic comparison does not cover. The first is a missing or stale
schedule join. Our join succeeds for over
$99\%$ of trips here, but that rate belongs to a curated challenge dataset, not
to a live feed. The second is frequency-based service, where no per-trip
scheduled time exists to difference against. The third is scoring a trip before
it finishes, since a deterministic median-delay comparison needs the completed
trip while a model reading telemetry prefixes does not. We test that
advantage directly in Section~\ref{sec:delay}.\ref{sec:early}.

\subsection{Contributions}

\begin{enumerate}[itemsep=0pt,parsep=1pt]
\item A complete, leakage-controlled pipeline from raw GPS to trip-level delay
      labels and trip-level anomaly scores, reporting how many trips each
      quality gate removes (Section~\ref{sec:prep}).
\item A direct comparison of latent representations against raw-feature
      baselines under identical folds, with paired significance tests
      (Section~\ref{sec:delay}). The result is negative for the latent
      representation and we report it as such.
\item An ablation showing that two of our three original anomaly validations
      were circular, driven by rainfall being simultaneously an AE input and the
      external validator (Section~\ref{sec:anom}).
\item A stratified analysis separating \emph{day type} from \emph{service
      disruption} in the day-level anomaly signal (Section~\ref{sec:anom}.\ref{sec:daytype}).
\item A study of alarm thresholds showing that the threshold rule limits
      detection far more than the score does
      (Section~\ref{sec:anom}.\ref{sec:threshold}).
\item An attribution analysis of what the delay model uses, and a
      probe showing that no individual latent dimension is interpretable
      (Section~\ref{sec:drivers}).
\item A network-level account of where the delays sit, including a negative
      result for passenger demand and a null result for trip-to-trip
      propagation (Section~\ref{sec:e10}).
\end{enumerate}

\section{Data}

We use three sources, all for the period 11--31 March 2026:

The \textbf{NetMob26 mobility feed}~\cite{netmob26} provides high-frequency
vehicle tracking, between $0.07$ and $1.13$ million GPS points per day (median
about $0.8$ million) and $13{,}850{,}467$ rows in total across the 19 dates
present. Dated GTFS schedule snapshots~\cite{gtfs} are used only to derive
delay labels.
\textbf{Hourly weather} from the public municipal feed supplies rainfall in
mm/h.

Two dates in the nominal range (18 and 19 March) are absent from the feed. A
further 23 trips carry a \texttt{service\_date} of 10 March, the result of
late-night trips whose service day rolls back past midnight; grouping is done on
\texttt{service\_date}, which therefore has 20 distinct values while the
calendar dates of trip starts number 19.

\section{Preprocessing}
\label{sec:prep}

\begin{figure*}[t]
\centering
\includegraphics[width=\textwidth]{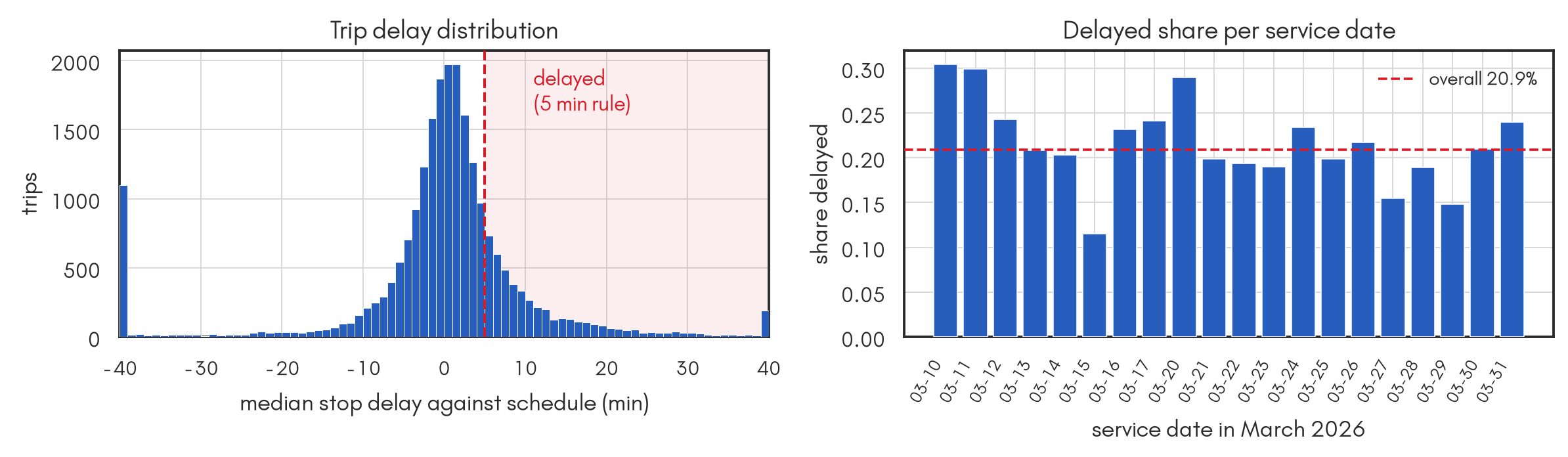}
\caption{How the delay label is made. Left, the distribution of median stop
delay against the schedule over the $22{,}605$ retained trips, with the
five-minute rule marked. The threshold falls on the shoulder of a broad
unimodal distribution rather than at a natural gap, so the delayed share
responds smoothly to where it is placed. The axis is clipped at $\pm 40$
minutes and the $5.6\%$ of trips beyond that are stacked into the end bins.
Right, the delayed share of each of the 20 service dates against the overall
rate of $20.9\%$. The day-to-day spread is what makes service date the
grouping variable for cross-validation
(Section~\ref{sec:method}.\ref{sec:cv}).}
\label{fig:label}
\end{figure*}

\subsection{Label derivation}

The dataset carries no delay annotations, so the label is derived. Each GPS trip
report is joined to its scheduled GTFS trip by matching identifiers per
direction against the dated snapshot, map-matched onto the route shape, and stop
arrival times are interpolated where along-route progress crosses each stop
position.

The Transit Capacity and Quality of Service Manual~\cite{tcqsm2013} defines a
fixed-route service as on-time when it runs no more than 5 minutes behind
schedule. Adopting this
industry boundary, we mark a trip \emph{delayed} when its median stop delay
against the schedule is at least 5 minutes, which labels $20.9\%$ of the
retained trips. Figure~\ref{fig:label} shows where that cut falls. It sits on
the flank of a single broad peak rather than in a gap, so the share is a
smooth function of the boundary rather than a stable feature of the data:
moving the rule to four minutes would label $25.2\%$ of trips and moving it to
six would label $17.7\%$. Nothing downstream depends on the exact position,
because every comparison in this report uses the same labels throughout.
We additionally require at least
five matched stops and $50\%$ stop coverage before trusting that median.
Coverage here is the share of scheduled stops between the first and last along-route positions reached by the GPS trace. It measures how much of the route was observed, not the accuracy of the positions.
A bus picked up late by the feed, or dropped early, scores low
even when every point it did report is precise.

\subsection{Quality gates}
\label{sec:gates}

Table~\ref{tab:gates} reports every gate with its exact threshold and the number
of candidate trip instances it removes.

\begin{table*}[t!]
\centering
\caption{Quality gates and how many trips each removes. Percentages are of the $38{,}411$
candidate trip instances. Gates are applied in the order listed; each count is
the number of instances that gate removed from what reached it.
$^{\dagger}$The multi-vehicle row is a deduplication rather than a rejection, since the
segment with the most points is kept and the others discarded, so the trip
survives. See Section~\ref{sec:prep}.\ref{sec:multiveh}. $^{\ddagger}$The
midnight-flip row is a reassignment: $835$ trips that crossed midnight were
moved to the adjacent service day and kept.}
\label{tab:gates}
\footnotesize
\tablefont
\begin{tabular}{llrr}
\toprule
Gate & Threshold & Dropped & \% \\
\midrule
No GTFS match            & no schedule join for trip id/direction   & $0$     & $0.00$ \\
Too few points           & $<10$ GPS points                          & $526$   & $1.37$ \\
Implausible duration     & outside $8$~min--$3$~h                    & $1{,}186$ & $3.09$ \\
Off-route                & $>40\%$ of points $>200$~m from shape     & $2{,}112$ & $5.50$ \\
Low stop coverage        & $<5$ matched stops or $<50\%$ coverage    & $6{,}609$ & $17.21$ \\
Multi-vehicle trip id$^{\dagger}$ & same trip id served by several vehicles   & $4{,}267$ & $11.11$ \\
Midnight flip$^{\ddagger}$ & trip reassigned to adjacent service day  & $0$     & $0.00$ \\
Day-letter mismatch      & calendar day vs service-pattern letter    & $0$     & $0.00$ \\
Schedule mismatch        & snapshot inconsistent with observed trip  & $1{,}106$ & $2.88$ \\
\midrule
\textbf{Retained}        &                                           & $\mathbf{22{,}605}$ & $\mathbf{58.9}$ \\
\bottomrule
\end{tabular}
\end{table*}

Of the retained $22{,}605$ trip instances across 20 service dates, $20.9\%$ are
labelled delayed. The complement matters as much, since $41.1\%$ of candidate
instances never reach the dataset. We examine the two largest
contributors below, the stop rule and the multi-vehicle deduplication, and both
turn out to be defensible. The remaining four, off-route removal
($5.5\%$), implausible duration ($3.1\%$), schedule mismatch ($2.9\%$) and
too-few-points ($1.4\%$), are applied on the
argument that each removes a trace we cannot interpret. We test them
collectively rather than one at a time in
Section~\ref{sec:delay}.\ref{sec:robust}, by relaxing every threshold at once
and rebuilding, and the conclusions do not depend on any of them. What that design
cannot tell us is how much each individual gate contributes.

Every one of these gates removes
traces that are short, broken or geometrically odd, and a genuinely disrupted
trip is more likely to produce exactly such a trace. If that were happening the
retained sample would be biased toward normal operation, which would matter most
for the anomaly analysis in Section~\ref{sec:anom}, since a detector evaluated
on a population with the worst trips filtered out looks better behaved than it
is.

The concern is testable in two ways that do not need delay labels for the
discarded trips. If the gates exclude disrupted operation, then days with
heavy filtering should look artificially punctual among the trips that survive.
They do not, since the Spearman rank correlation~\cite{Spearman1904} between a
day's drop rate and the delay rate of its retained trips is $\rho = +0.668$
($p = 0.002$), the opposite sign to the one the concern predicts. Nor do the known degraded-telemetry days lose more trips
than normal days ($42.8\%$ against $40.4\%$, $p = 0.76$). The second test
uses labels we do have. The relaxed rebuild of
Section~\ref{sec:prep}.\ref{sec:gatecheck} recovers $4{,}889$ trips the stop
rule had discarded, and those trips are delayed $17.8\%$ of the time against
$20.9\%$ for the trips that were kept. On both tests the discarded population
is, if anything, less disrupted than the retained one.

One residual signal runs the other way, since days with higher drop rates
carry slightly lower median anomaly scores among their surviving trips
($\rho = -0.507$, $p = 0.027$), which is the pattern filtering-out-the-unusual
would produce. We are reluctant to read much into it,
because the anomaly score is itself dominated by day type
(Section~\ref{sec:anom}.\ref{sec:daytype}) and the drop rate tracks telemetry
volume, so the two are confounded through the calendar. The delay-side evidence
is the cleaner of the two and it does not support the concern.

The dominant gate is stop coverage ($17.2\%$), followed by the multi-vehicle
deduplication ($11.1\%$ of instances). The first removes trips whose delay
median would not be trustworthy, the second removes duplicate views of a trip
rather than the trip itself, and neither selects on operating condition, as the
tests above and Section~\ref{sec:prep}.\ref{sec:multiveh} show.

\subsection{Sequence construction}

Each trip becomes a $64 \times 12$ array of 1-minute bins. The bin width matches
the telemetry, which reports every 30 to 60 seconds, so one minute is the
finest grid that does not mostly interpolate between observations. The window of
64 bins is a compromise, since trips have a median
duration of $86$ minutes and only $24.6\%$ finish inside 64 minutes, so three
quarters of trips are truncated and the model sees their first hour rather than
the whole route. We accept that because the alternative, a window long enough
for the $90$th percentile at $141$ minutes, would more than double the sequence
length and leave most cells padded --- and because
Section~\ref{sec:delay}.\ref{sec:early} shows performance is already near its
ceiling at a quarter of the route, so the discarded tail carries little
additional signal. The 12 channels are the quantities derivable from GPS,
weather and network structure that do not encode the label. Three are kinematic (speed, acceleration and heading change) and three describe
route geometry (progress fraction, progress rate and cross-track error). One
records dwell, the fraction of the bin spent below $0.5$~m/s. Four carry
context: time of day as a sine and cosine pair, a weekend flag and hourly
rainfall. The last is a network term, the headway to the preceding trip of the
same line and direction, capped at 60 minutes.

No schedule-derived quantity enters the features, so the delay label cannot leak
into the representation.

\textbf{Padding and masking.} Trips shorter than 64 minutes are zero-padded and
carry an explicit length, so both the pooled latent and the reconstruction error
are masked to real timesteps. Padding contributes nothing to either.

\textbf{Missing values.} Empty bins are filled from the interpolated along-route
progress; cross-track error is carried forward; rainfall is forward-filled from
the hourly feed.

\textbf{Task framing.} Each trip is scored after it has finished, from the first
64 minutes of its trajectory. The task as posed is therefore
\emph{retrospective classification} rather than early prediction.

\subsection{Sensitivity to the stop-coverage rule}
\label{sec:gatecheck}

The stop rule discards more trips than any other gate, so we test it directly.
Splitting the two clauses apart shows the work
is done almost entirely by one of them. Of the $6{,}609$ removals, $3{,}710$
($9.66\%$ of candidates) fail only the $50\%$ coverage test and $2{,}797$
($7.28\%$) fail both, while just $102$ trips ($0.27\%$) fail only the five-stop
floor. The rule is $41.8\%$ of all rejections in the pipeline. Surviving trips
are not crowded against the threshold, with median coverage $0.931$, and
only $2.4\%$ of them sit between $0.50$ and $0.55$.

To test whether the discarded trips carry usable labels we rebuilt the dataset
with the rule relaxed to a single matched stop, recovering $4{,}889$ trips for a
total of $27{,}494$, and scored the two populations separately
(Table~\ref{tab:covsens}). The recovered trips are exactly the marginal ones, with median
coverage $0.250$ against $0.931$ for the trips already kept, and a median of
three matched stops in the lowest coverage band.

\begin{table}[htbp]
\centering
\caption{RAW-STATS+XGBoost under the relaxed rule, mean over five folds.
Training on the recovered trips is harmless; scoring on them is not.}
\label{tab:covsens}
\footnotesize
\setlength{\tabcolsep}{4pt}
\tablefont
\begin{tabular}{lcc}
\toprule
Train / test split & ROC-AUC & PR-AUC \\
\midrule
strict train, strict test      & $0.850$ & $0.648$ \\
strict train, recovered test   & $0.785$ & $0.522$ \\
relaxed train, strict test     & $0.848$ & $0.643$ \\
relaxed train, all test        & $0.853$ & $0.641$ \\
\bottomrule
\end{tabular}
\end{table}

A model trained on strict data loses a lot of ground on the recovered trips,
falling from $0.850$ to $0.785$ ROC-AUC and from $0.648$ to $0.522$ PR-AUC,
which is what a median taken over three stops should look like. Yet adding
those trips to training changes nothing on clean test data ($0.848$ against
$0.850$), so the gate is protecting the labels rather than the model. The
headline numbers are therefore not an artefact of the gate; relaxing it would
mainly add noise to the evaluation. Thus, we keep the rule.

On the relaxed data, scored by a weather-free CNN-AE retrained there,
the anomaly results survive the same treatment. The weekend effect, the degraded-day
separation within both day strata, and the ranking of threshold rules all
reproduce (per-line calibration again lifts the 22 March alarm rate most, to
$25.4\%$), and the day-level ordering matches the strict one at Spearman
$\rho = 0.991$. The rain ablation reproduces almost exactly, since with
rainfall in the input $11.7\%$ of rain-hour trips are flagged against $4.4\%$
dry, and retraining without it flattens this to $5.1\%$ against $5.0\%$
($p = 0.97$), the same collapse reported in
Section~\ref{sec:anom}.\ref{sec:circular} on the strict data. The chronic
line-directions of Section~\ref{sec:e10} also hold, with $24$ pairs carrying
$57.7\%$ of delayed trips.

That test covers one model, while the claim this report rests on is the
comparison between representations, so we retrained the CNN-AE on the relaxed
data as well and repeated it. Raw features win by a wider margin there, with
RAW-STATS+XGBoost reaching $0.853 \pm 0.011$ ROC-AUC and $0.641 \pm 0.025$
PR-AUC against $0.839 \pm 0.016$ and $0.623 \pm 0.023$ for CNN-AE+XGBoost, a
paired difference of $-0.0141$ ($p = 0.028$) and $-0.0173$ ($p = 0.033$). On the
strict dataset neither gap was significant ($p = 0.213$ ROC-AUC); with the
discarded trips restored it is. The central negative result of this report is
therefore not an artefact of the stop rule, and if anything the rule was
flattering to the autoencoder.

\subsection{Trips served by more than one vehicle}
\label{sec:multiveh}

The multi-vehicle counter in Table~\ref{tab:gates} is not a rejection like the
other gates. When a \texttt{tripId} appears under several
vehicle identifiers, the pipeline keeps the segment with the most GPS points and
drops the rest, so the trip survives and what is lost is every other vehicle's
view of it. The $4{,}267$ figure counts superseded \emph{instances}, not
discarded trips.

Across the period, $2{,}456$ trip identifiers are served by more than one
vehicle, usually two ($2{,}139$ cases) or three ($244$), with a small tail of
identifiers shared by many vehicles, up to $73$. Most are not duplicated
identifiers, because in $88\%$ of cases the segments do not
overlap in time at all, which is the signature of a genuine vehicle change mid-service
rather than a data fault. Only $2.2\%$ overlap substantially.

The cost is that the kept segment is often a minority of the trip. Measured
against the span covered by all vehicles together, the retained segment has a
median ratio of $0.63$, falls below $0.80$ for $75.9\%$ of these trips and
below $0.50$ for $31.1\%$. In total $71.4\%$ of the GPS points recorded on
these trips are thrown away. The effect is concentrated, with line 45 alone accounting
for $13.9\%$ of the cases.

This matters for the delay label, because these trips are visibly different.
Among retained trips, those touched by the deduplication are delayed $28.6\%$
of the time against $18.9\%$ for the rest. Their mean coverage is barely lower
($0.879$ against $0.899$), but that comparison is conditioned on survival, because the
trips whose kept segment was too short were already removed by the stop rule.
So deduplication is not what drives the coverage loss, yet a fifth of the
dataset is still built from partial trajectories whose missing portion was
recorded under another identifier.

We therefore checked whether this biases the results, since discarding $71.4\%$
of the points on a fifth of the data is not obviously harmless. It does not move
the headline. Excluding every affected trip from both training and test changes
RAW-STATS+XGBoost from $0.846$ to $0.853 \pm 0.025$ ROC-AUC, inside the
cross-fold spread, so no conclusion in this report depends on them. Per trip,
though, the model is measurably worse on them, at $0.810$ ROC-AUC against $0.854$
on single-vehicle trips. Comparisons of PR-AUC between the two groups are not
meaningful, because their delayed prevalence differs ($0.283$ against $0.190$)
and PR-AUC rises with prevalence by construction.

Truncated trips sit lower in \texttt{mean(progress\_frac)} than untruncated
ones at the same label, by $0.043$ among on-time trips, and lower is the
direction that predicts delay, which is the mechanism behind that gap.
Against a label separation of roughly $0.14$ in that feature, the truncation
shift is around a third as large, which is enough to push truncated on-time
trips toward false positives.

The elevated delay rate on these trips admits two explanations. Either it is real, with buses relieved mid-service because they are already running late, or it is an artefact of taking the median over a partial trajectory. Both are consistent with the observed rate, so we distinguished them by rebuilding the dataset with the segments stitched together in time order rather than discarded, which is admissible because $88\%$ of these cases are sequential hand-overs with no temporal overlap. The merged dataset contains $22{,}417$ trips with a delayed share of $20.5\%$. Among the affected trips, the delay rate falls from $28.6\%$ to $26.8\%$, while the rate for the remainder is unchanged at $18.9\%$. Truncation therefore accounts for under a fifth of the gap; the remainder is attributable to the trips themselves. This is consistent with the interpretation that buses already running late are more likely to be relieved mid-service.

\section{Method}
\label{sec:method}

\subsection{Autoencoders}

Two AE variants are compared under one scaffold, both with a 64-dimensional
latent space. The \textbf{CNN-AE} is a one-dimensional convolutional~\cite{LeCun1998cnn}
autoencoder~\cite{Hinton2006autoencoder}. Its encoder stacks 3 Conv1d blocks
(kernel size 5, ``same'' padding) over the time axis, then a $1\times1$
convolution to a per-timestep latent $z_{\mathrm{seq}}$, with length-masked mean
pooling down to $z$. The decoder broadcasts $z$ over $T$ with a learned
positional embedding. The \textbf{LSTM-AE} is a sequence-to-sequence
recurrent~\cite{Hochreiter1997lstm} autoencoder, as applied to bus trajectories
by Zhang et al.~\cite{Zhang2021bus}.

Both are trained for 20 epochs, batch size 64, learning rate $10^{-3}$, with
Adam~\cite{Kingma2015adam}.

\subsection{Downstream classifiers}

For each fold the AE is trained unsupervised on the training split only. The
frozen encoder then produces a pooled latent $z$ for four classifiers
(MLP~\cite{Rumelhart1986}, XGBoost~\cite{Chen2016xgboost}, random forest~\cite{Breiman2001rf}, logistic regression) and a latent sequence
$z_{\mathrm{seq}}$ for two recurrent classifiers (LSTM~\cite{Hochreiter1997lstm},
GRU~\cite{Cho2014gru}).

\subsection{Cross-validation}
\label{sec:cv}

Evaluation uses \texttt{StratifiedGroupKFold}~\cite{Pedregosa2011sklearn} with 5 folds grouped by
\emph{service date}. Day-level effects (weather, traffic, telemetry quality, and
as Section~\ref{sec:anom}.\ref{sec:daytype} shows, day type) are the main leakage
channel, so every model is scored only on unseen days.

\textbf{Metrics.} We report ROC-AUC~\cite{Fawcett2006roc} and the area under
the precision--recall curve, computed as average precision. At a delayed-class prevalence of $20.9\%$ the two
can disagree, and PR-AUC is the more informative of the pair under class
imbalance~\cite{Davis2006prroc}, so we lead with it wherever they conflict.
Paired comparisons over the five folds use the $t$-test~\cite{Student1908}
alongside the Wilcoxon
signed-rank test~\cite{Wilcoxon1945}; group comparisons that do not assume
normality use the Mann--Whitney $U$ test~\cite{Mann1947} and rank correlations
use Spearman's $\rho$~\cite{Spearman1904}.

\textbf{Hyperparameters.} No systematic search was performed. All models use
fixed settings taken from their respective literature. Section~\ref{sec:delay}.\ref{sec:tuning} tests whether
tuning changes any of the conclusions.

\subsection{Anomaly score}
\label{sec:anomscore}

Each trip is scored by the AE of the fold in which it was held out, so no trip
is scored by a model that saw it during training.

\begin{figure*}[t!]
\centering
\includegraphics[width=\textwidth]{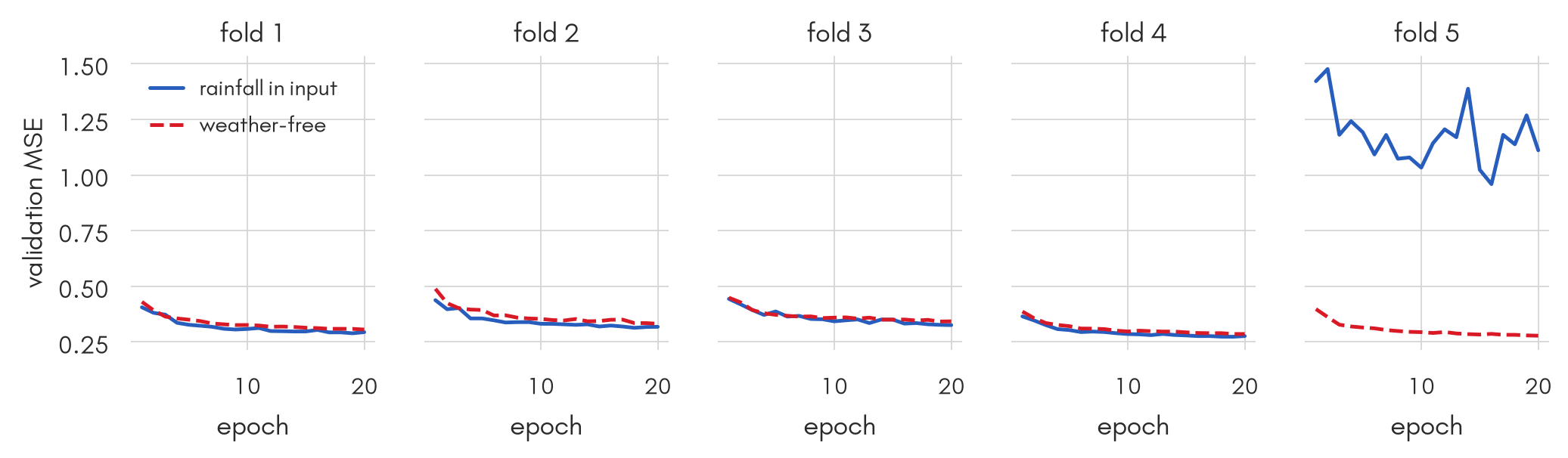}
\caption{Validation reconstruction MSE per fold, for the CNN-AE with rainfall in
the input (solid) and without it (dashed). Fold~5 holds out 23 March, the
11.8~mm/h rain day. With rainfall as an input its error runs roughly four times
higher than any other fold and never settles; drop the channel and the fold
behaves like the rest. Error scales differ enough between folds that pooling raw
MSE would confound fold identity with anomaly, which is why we standardise
within each fold.}
\label{fig:recon}
\end{figure*}

Because every fold has its own AE, raw reconstruction errors are not comparable
across folds. The heavy-tailed MSE is therefore log-transformed and standardised
\emph{within} each fold before pooling; the threshold is applied once to the
pooled score. Figure~\ref{fig:recon} shows the per-fold reconstruction curves
behind this step. The autoencoders settle on similar, though not identical,
error scales, with one clear exception. Fold~5 holds out 23 March, the heaviest
rain of the period, and its validation error stays roughly four times higher
than the other folds for the whole run ($1.11$ against $0.28$ to $0.33$ at the
last epoch). Retraining without the rainfall channel brings it back into line at
$0.28$. The autoencoder cannot reconstruct rare rainfall values, the same
mechanism that Section~\ref{sec:anom}.\ref{sec:circular} shows contaminating the
anomaly score.

\section{Delay detection results}
\label{sec:delay}

\subsection{Comparison of representations}

\begin{table*}[t!]
\centering
\caption{Delay detection on unseen days: mean $\pm$ SD over the five
day-grouped folds. Delayed-class prevalence is $0.209$. F1 is binary F1 on the
delayed (positive) class. Best value per column in bold.}
\label{tab:full}
\scriptsize
\tablefont
\begin{tabular}{llccccc}
\toprule
Representation & Classifier & ROC-AUC & PR-AUC & Precision & Recall & F1 \\
\midrule
CNN-AE  & MLP      & $0.824 \pm 0.021$ & $0.593 \pm 0.054$ & $0.437 \pm 0.032$ & $0.738 \pm 0.072$ & $0.548 \pm 0.037$ \\
CNN-AE  & XGBoost  & $0.838 \pm 0.016$ & $0.629 \pm 0.051$ & $0.539 \pm 0.056$ & $0.637 \pm 0.054$ & $0.581 \pm 0.039$ \\
CNN-AE  & RF       & $0.839 \pm 0.015$ & $0.621 \pm 0.053$ & $0.546 \pm 0.057$ & $0.612 \pm 0.062$ & $0.574 \pm 0.039$ \\
CNN-AE  & LR       & $0.811 \pm 0.017$ & $0.564 \pm 0.054$ & $0.446 \pm 0.043$ & $0.694 \pm 0.057$ & $0.541 \pm 0.040$ \\
CNN-AE  & LSTM     & $0.826 \pm 0.020$ & $0.601 \pm 0.052$ & $0.448 \pm 0.040$ & $0.738 \pm 0.058$ & $0.557 \pm 0.041$ \\
CNN-AE  & GRU      & $0.828 \pm 0.023$ & $0.607 \pm 0.052$ & $0.437 \pm 0.050$ & $\mathbf{0.752 \pm 0.048}$ & $0.552 \pm 0.048$ \\
\midrule
LSTM-AE & MLP      & $0.801 \pm 0.019$ & $0.548 \pm 0.067$ & $0.421 \pm 0.053$ & $0.698 \pm 0.085$ & $0.522 \pm 0.048$ \\
LSTM-AE & XGBoost  & $0.814 \pm 0.016$ & $0.575 \pm 0.062$ & $0.471 \pm 0.053$ & $0.650 \pm 0.054$ & $0.544 \pm 0.044$ \\
LSTM-AE & RF       & $0.819 \pm 0.017$ & $0.576 \pm 0.061$ & $0.476 \pm 0.055$ & $0.645 \pm 0.064$ & $0.545 \pm 0.047$ \\
LSTM-AE & LR       & $0.785 \pm 0.016$ & $0.512 \pm 0.057$ & $0.403 \pm 0.043$ & $0.703 \pm 0.048$ & $0.512 \pm 0.045$ \\
LSTM-AE & LSTM     & $0.809 \pm 0.020$ & $0.559 \pm 0.068$ & $0.418 \pm 0.048$ & $0.717 \pm 0.088$ & $0.526 \pm 0.047$ \\
LSTM-AE & GRU      & $0.808 \pm 0.018$ & $0.554 \pm 0.059$ & $0.407 \pm 0.055$ & $0.751 \pm 0.059$ & $0.525 \pm 0.048$ \\
\midrule
RAW-STATS & XGBoost & $\mathbf{0.846 \pm 0.013}$ & $\mathbf{0.637 \pm 0.053}$ & $0.537 \pm 0.054$ & $0.655 \pm 0.061$ & $\mathbf{0.587 \pm 0.039}$ \\
RAW-STATS & RF      & $0.834 \pm 0.018$ & $0.610 \pm 0.063$ & $0.525 \pm 0.065$ & $0.630 \pm 0.072$ & $0.569 \pm 0.048$ \\
RAW-STATS & LR      & $0.803 \pm 0.012$ & $0.531 \pm 0.048$ & $0.436 \pm 0.037$ & $0.690 \pm 0.054$ & $0.534 \pm 0.039$ \\
RAW-FLAT  & XGBoost & $0.845 \pm 0.015$ & $0.632 \pm 0.053$ & $\mathbf{0.568 \pm 0.050}$ & $0.606 \pm 0.056$ & $0.585 \pm 0.046$ \\
RAW-FLAT  & RF      & $0.842 \pm 0.018$ & $0.622 \pm 0.058$ & $0.537 \pm 0.060$ & $0.632 \pm 0.056$ & $0.578 \pm 0.044$ \\
RAW-FLAT  & LR      & $0.784 \pm 0.017$ & $0.500 \pm 0.059$ & $0.424 \pm 0.041$ & $0.674 \pm 0.040$ & $0.520 \pm 0.038$ \\
\midrule
SimCLR  & MLP      & $0.824 \pm 0.015$ & $0.596 \pm 0.054$ & $0.424 \pm 0.043$ & $0.747 \pm 0.049$ & $0.541 \pm 0.045$ \\
SimCLR  & XGBoost  & $0.819 \pm 0.022$ & $0.595 \pm 0.063$ & $0.504 \pm 0.058$ & $0.609 \pm 0.052$ & $0.550 \pm 0.048$ \\
SimCLR  & RF       & $0.815 \pm 0.023$ & $0.582 \pm 0.068$ & $0.508 \pm 0.046$ & $0.591 \pm 0.072$ & $0.544 \pm 0.047$ \\
SimCLR  & LR       & $0.807 \pm 0.016$ & $0.556 \pm 0.057$ & $0.441 \pm 0.054$ & $0.688 \pm 0.050$ & $0.535 \pm 0.047$ \\
\bottomrule
\end{tabular}
\end{table*}

Because this comparison tests the premise of the work, we run it over five
representations under identical folds. The first two are the proposed pipeline, the
\textbf{CNN-AE} and \textbf{LSTM-AE} latents. \textbf{RAW-STATS} is a plain
tabular baseline, the per-trip mean, standard deviation, minimum and maximum of
each of the 12 channels over the real timesteps, 48 features in all.
\textbf{RAW-FLAT} feeds the
flattened $64\times12 = 768$-dimensional raw sequence straight to the tabular
models. \textbf{SimCLR}~\cite{Chen2020simclr} is a contrastive encoder trained on the
same folds, a learned representation with no reconstruction objective.

Table~\ref{tab:full} reports all metrics for all configurations.

\subsection{Paired significance tests}

Differences between the best configurations are small relative to cross-fold
variability, so point estimates alone cannot support a claim of superiority.
Table~\ref{tab:paired} reports paired tests over the five folds. With $n = 5$,
the Wilcoxon signed-rank test has a minimum attainable $p$ of $0.031$; we report
both it and the paired $t$-test.

\begin{table*}[t!]
\centering
\caption{Paired comparisons over the five folds. Positive difference favours the
first model. $p_t$ is the paired $t$-test, $p_W$ the Wilcoxon signed-rank test.}
\label{tab:paired}
\footnotesize
\tablefont
\begin{tabular}{llrrr}
\toprule
Comparison & Metric & Difference & $p_t$ & $p_W$ \\
\midrule
CNN-AE+XGBoost vs RAW-STATS+XGBoost & ROC-AUC & $-0.0076$ & $0.213$ & $0.188$ \\
CNN-AE+XGBoost vs RAW-STATS+XGBoost & PR-AUC  & $-0.0082$ & $0.087$ & $0.062$ \\
CNN-AE+XGBoost vs RAW-FLAT+XGBoost  & ROC-AUC & $-0.0064$ & $0.264$ & $0.312$ \\
CNN-AE+XGBoost vs RAW-FLAT+XGBoost  & PR-AUC  & $-0.0027$ & $0.510$ & $0.438$ \\
\midrule
CNN-AE+RF vs RAW-STATS+RF           & ROC-AUC & $+0.0053$ & $0.365$ & $0.625$ \\
CNN-AE+RF vs RAW-STATS+RF           & PR-AUC  & $+0.0108$ & $0.262$ & $0.312$ \\
\midrule
CNN-AE+LR vs RAW-STATS+LR           & ROC-AUC & $+0.0079$ & $0.244$ & $0.312$ \\
CNN-AE+LR vs RAW-STATS+LR           & PR-AUC  & $+0.0332$ & $\mathbf{0.001}$ & $0.062$ \\
\midrule
CNN-AE+XGBoost vs SimCLR+XGBoost    & ROC-AUC & $+0.0189$ & $0.072$ & $0.125$ \\
CNN-AE+XGBoost vs SimCLR+XGBoost    & PR-AUC  & $+0.0340$ & $\mathbf{0.017}$ & $0.062$ \\
\midrule
CNN-AE+XGBoost vs LSTM-AE+XGBoost   & ROC-AUC & $+0.0242$ & $\mathbf{0.007}$ & $0.062$ \\
CNN-AE+RF vs CNN-AE+XGBoost         & ROC-AUC & $+0.0008$ & $0.724$ & $0.812$ \\
\bottomrule
\end{tabular}
\end{table*}

\subsection{Interpretation}

\textbf{The latent representation does not improve accuracy.} RAW-STATS+XGBoost
is the best configuration overall on ROC-AUC, PR-AUC and F1. Against the best
latent configuration neither gap reaches significance on this dataset
($p_t = 0.213$ ROC-AUC, $p_t = 0.087$ PR-AUC), though PR-AUC, the more
informative metric at $20.9\%$ prevalence, comes closer. The separation does
become significant once the preprocessing filters are relaxed and the dataset
grows (Section~\ref{sec:delay}.\ref{sec:robust}), which is where that claim
rests. A simple per-trip summary of the same
12 channels is at least as good as, and on the precision--recall trade-off
slightly better than, a learned 64-dimensional embedding.

\textbf{The latent space does buy linear separability.} Logistic regression
gains $+0.033$ PR-AUC over RAW-STATS ($p_t = 0.001$) and $+0.064$ over
RAW-FLAT, the weaker of the two raw baselines, against ROC-AUC gains of only
$+0.008$ and $+0.027$.
This is the clearest benefit we can defend. The autoencoder rearranges the trip
into a space where a straight-line decision boundary works far better than it
does on the raw channels. Whether that matters in practice depends on whether a
linear model is required, whether for interpretability, for speed, or for
running on small hardware.

\textbf{Reconstruction beats contrastive, and convolution beats recurrence.}
CNN-AE+XGBoost beats SimCLR+XGBoost by $+0.019$ ROC-AUC ($p_t = 0.072$) and
$+0.034$ PR-AUC ($p_t = 0.017$), so the PR-AUC advantage holds while the
ROC-AUC one does not reach significance. It beats LSTM-AE+XGBoost by $+0.024$
ROC-AUC ($p_t = 0.007$). The same CNN-over-LSTM ordering holds for every
classifier in Table~\ref{tab:full}, but the paired test was run only on the
best pair. Within the CNN-AE, random forest and XGBoost are
indistinguishable ($p_t = 0.724$).

\subsection{Effect of hyperparameter tuning}
\label{sec:tuning}

Every number above uses literature defaults, so the comparison between
representations is a comparison of untuned models. We ran a grid search with
selection nested inside each fold, so candidates are scored on an inner split
taken from that fold's training days only, the winner is refit on the full
training split, and only then is it scored on the outer test fold, so no
configuration sees a test day before being chosen.

Tuning helps, but not enough to disturb the conclusion. Random forest gains the
most ($+0.008$ ROC-AUC, $+0.015$ PR-AUC, both $p < 0.01$) and XGBoost gains
little ($+0.002$ and $+0.007$, $p = 0.171$ and $p = 0.094$). Logistic regression
moves in opposite directions on the two metrics, $-0.0013$ ROC-AUC
($p = 0.032$) and $+0.0022$ PR-AUC ($p = 0.023$); both clear $p < 0.05$ because
the per-fold differences are extremely consistent, but the magnitudes are two
orders below anything that would change a decision. The tuned RAW-STATS+XGBoost reaches $0.848$ ROC-AUC, still above the
best latent configuration's untuned $0.838$, so the ordering between raw
features and latent features is unaffected. The search was run on the
classifier side only; the AE architecture itself was not searched, and that
remains a limitation.

\subsection{Robustness to preprocessing and architecture}
\label{sec:robust}

The comparison above rests on one dataset and one autoencoder configuration, so
we varied both.

Relaxing the preprocessing does not rescue the latent representation; it widens
the gap (Table~\ref{tab:robust}). Putting back the trips the stop rule discards gives a difference of
$-0.014$ ROC-AUC ($p = 0.028$), stitching the multi-vehicle segments gives
$-0.010$ ($p = 0.155$), and loosening every gate at once, which recovers
$7{,}941$ extra trips for a total of $30{,}546$, gives $-0.024$ ($p = 0.067$)
with PR-AUC at $-0.031$ ($p = 0.040$). The direction is the same in all four
datasets and the magnitude grows as filtering relaxes, while the autoencoder's
cross-fold spread widens from $\pm 0.016$ to $\pm 0.042$ and the raw baseline
stays steady. The delayed share barely moves across all of this, $20.91\%$ on
the fully relaxed data against $20.92\%$ on the strict data, which is further
evidence that the filters are not selecting on the label.

Nor is the gap an artefact of the architecture we happened to pick
(Table~\ref{tab:arch}). Sweeping
latent dimension over $16$, $32$, $64$ and $128$ and depth over two, three and
four convolutional blocks, the best configuration reaches $0.8377$ ROC-AUC,
still short of the raw baseline's $0.8459$. The default used throughout this
report, $64$ dimensions and three blocks, scores $0.8375$, within $0.0002$ of
the best setting; doubling the latent space to $128$ adds nothing, cutting it to
$16$ costs $0.011$, and a fourth block hurts. We report this as a sweep rather
than a nested selection on purpose, because the useful statement is not which
setting wins but that none of them reaches the baseline, which holds whatever
protocol would have chosen among them.

\begin{table*}[t!]
\centering
\caption{Raw features against the latent representation across four versions of
the dataset. Each row is a full rebuild of the preprocessing followed by a fresh
autoencoder and a fresh baseline on the same day-grouped folds. Differences are
CNN-AE minus RAW-STATS, so negative favours the raw features; $p_t$ is a paired
$t$-test over the five folds.}
\label{tab:robust}
\footnotesize
\setlength{\tabcolsep}{4pt}
\tablefont
\begin{tabular}{lrrcccccc}
\toprule
& & & \multicolumn{3}{c}{ROC-AUC} & \multicolumn{3}{c}{PR-AUC} \\
\cmidrule(lr){4-6}\cmidrule(lr){7-9}
Dataset & Trips & Delayed & RAW-STATS & CNN-AE & $\Delta$ ($p_t$) & RAW-STATS & CNN-AE & $\Delta$ ($p_t$) \\
\midrule
strict (reported) & $22{,}605$ & $20.92\%$
  & $0.846 \pm 0.013$ & $0.838 \pm 0.016$ & $-0.008$ ($0.213$)
  & $0.637 \pm 0.053$ & $0.629 \pm 0.051$ & $-0.008$ ($0.087$) \\
stop rule relaxed & $27{,}494$ & $20.36\%$
  & $0.853 \pm 0.011$ & $0.839 \pm 0.016$ & $-0.014$ ($\mathbf{0.028}$)
  & $0.641 \pm 0.025$ & $0.623 \pm 0.023$ & $-0.017$ ($\mathbf{0.033}$) \\
multi-vehicle merged & $22{,}417$ & $20.52\%$
  & $0.847 \pm 0.015$ & $0.837 \pm 0.022$ & $-0.010$ ($0.155$)
  & $0.639 \pm 0.053$ & $0.627 \pm 0.064$ & $-0.013$ ($0.191$) \\
all gates relaxed & $30{,}546$ & $20.91\%$
  & $0.854 \pm 0.023$ & $0.830 \pm 0.042$ & $-0.024$ ($0.067$)
  & $0.652 \pm 0.055$ & $0.620 \pm 0.074$ & $-0.031$ ($\mathbf{0.040}$) \\
\bottomrule
\end{tabular}
\end{table*}

\begin{table}[t]
\centering
\caption{Autoencoder architecture sweep, CNN-AE+XGBoost, mean over the five
folds, ordered by descending latent dimension then depth. No configuration reaches the
raw-feature baseline.}
\label{tab:arch}
\footnotesize
\setlength{\tabcolsep}{5pt}
\tablefont
\begin{tabular}{rrcc}
\toprule
Latent dim & Conv blocks & ROC-AUC & PR-AUC \\
\midrule
$128$ & $3$ & $0.8377$ & $0.6326$ \\
$64$  & $4$ & $0.8338$ & $0.6198$ \\
$64$  & $3$ & $0.8375$ & $0.6292$ \\
$64$  & $2$ & $0.8366$ & $0.6298$ \\
$32$  & $3$ & $0.8344$ & $0.6237$ \\
$16$  & $3$ & $0.8262$ & $0.6078$ \\
\midrule
\multicolumn{2}{l}{\textbf{RAW-STATS+XGBoost}} & $\mathbf{0.8459}$ & $\mathbf{0.6370}$ \\
\bottomrule
\end{tabular}

\vspace{2pt}
{\footnotesize The row in bold is not an autoencoder. $64\times3$ is the
configuration used throughout this report.}
\end{table}

\subsection{Latent space structure}

\begin{figure*}[t]
\centering
\includegraphics[width=0.92\textwidth]{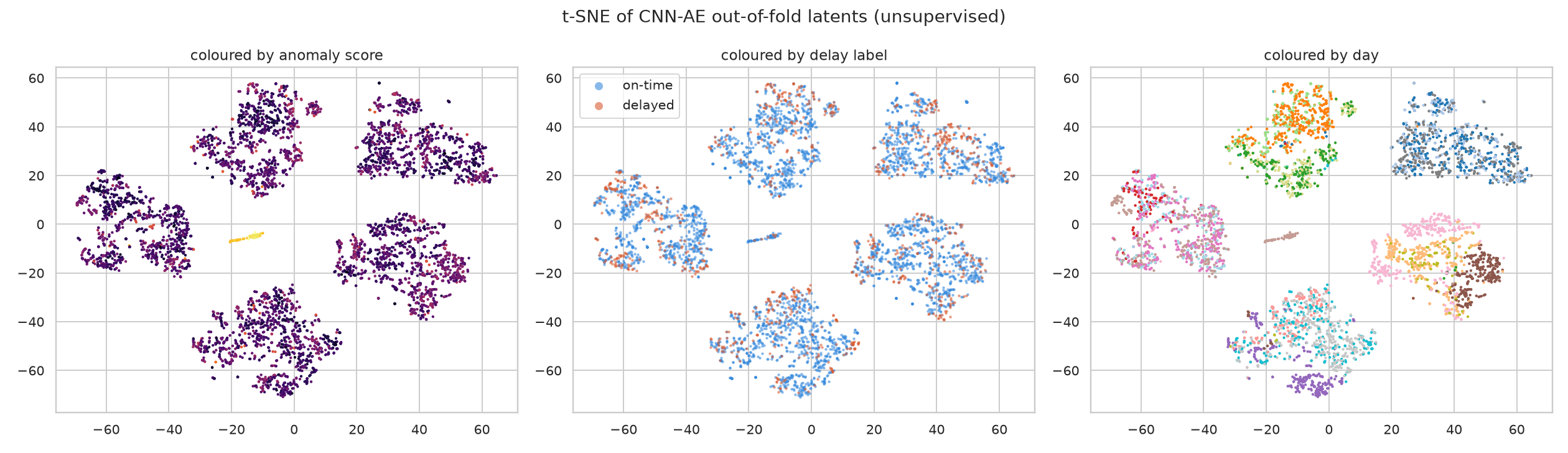}
\caption{t-SNE of $5{,}000$ out-of-fold CNN-AE latents, coloured by anomaly
score (left), delay label (centre) and service date (right). The five lobes
correspond to the five fold-specific encoders.}
\label{fig:tsne}
\end{figure*}

Figure~\ref{fig:tsne} shows a t-SNE~\cite{Maaten2008tsne} projection of
out-of-fold CNN-AE latents.
The five visible lobes are an artefact of pooling five fold-specific encoders
rather than structure in the data, and a single globally trained AE
(Section~\ref{sec:anom}.\ref{sec:globalae}) cannot produce them.
Within each lobe, delayed trips concentrate at distinct margins rather than
scattering uniformly, and trips of the same service date cluster together.

\subsection{Early prediction from trip prefixes}
\label{sec:early}

Section~\ref{sec:intro}.\ref{sec:why} argues that a learned model is useful
partly because it can score a trip that is still running, where the
deterministic schedule comparison needs the completed trip. We test it by
truncating every trip to the first
$25\%$, $50\%$ and $75\%$ of its own real length, zero the remainder and
shorten the recorded length, then rerun the whole comparison on the same
day-grouped folds. Table~\ref{tab:early} reports the result.

\begin{table}[b]
\centering
\caption{Delay detection from a prefix of the trip, RAW-STATS+XGBoost, mean
$\pm$ SD over the five folds. The last column is the share of the full-trip
discriminative power retained, measured above the $0.5$ chance baseline.}
\label{tab:early}
\footnotesize
\setlength{\tabcolsep}{5pt}
\tablefont
\begin{tabular}{lccc}
\toprule
Route seen & ROC-AUC & PR-AUC & Skill retained \\
\midrule
$25\%$  & $0.824 \pm 0.011$ & $0.581 \pm 0.046$ & $93.8\%$ \\
$50\%$  & $0.838 \pm 0.012$ & $0.620 \pm 0.048$ & $97.7\%$ \\
$75\%$  & $0.844 \pm 0.012$ & $0.635 \pm 0.051$ & $99.6\%$ \\
$100\%$ & $0.846 \pm 0.013$ & $0.637 \pm 0.053$ & $100\%$ \\
\bottomrule
\end{tabular}
\end{table}

At a quarter of the route the model already holds $93.8\%$ of the discriminative
power it reaches with the complete trajectory, and at half the route $97.7\%$,
so most of the signal is available early. The schedule comparison can say
nothing at either point, because the median stop delay it needs is not defined
until the trip ends. This is the strongest operational case in the report for
a model that reads telemetry, here the raw-feature baseline, and it agrees with
Section~\ref{sec:drivers}, where the dominant features are averages of speed
and along-route progress, which stabilise well before a trip finishes.

\section{Feature attribution}
\label{sec:drivers}

\begin{figure*}[t]
\centering
\includegraphics[width=0.92\textwidth]{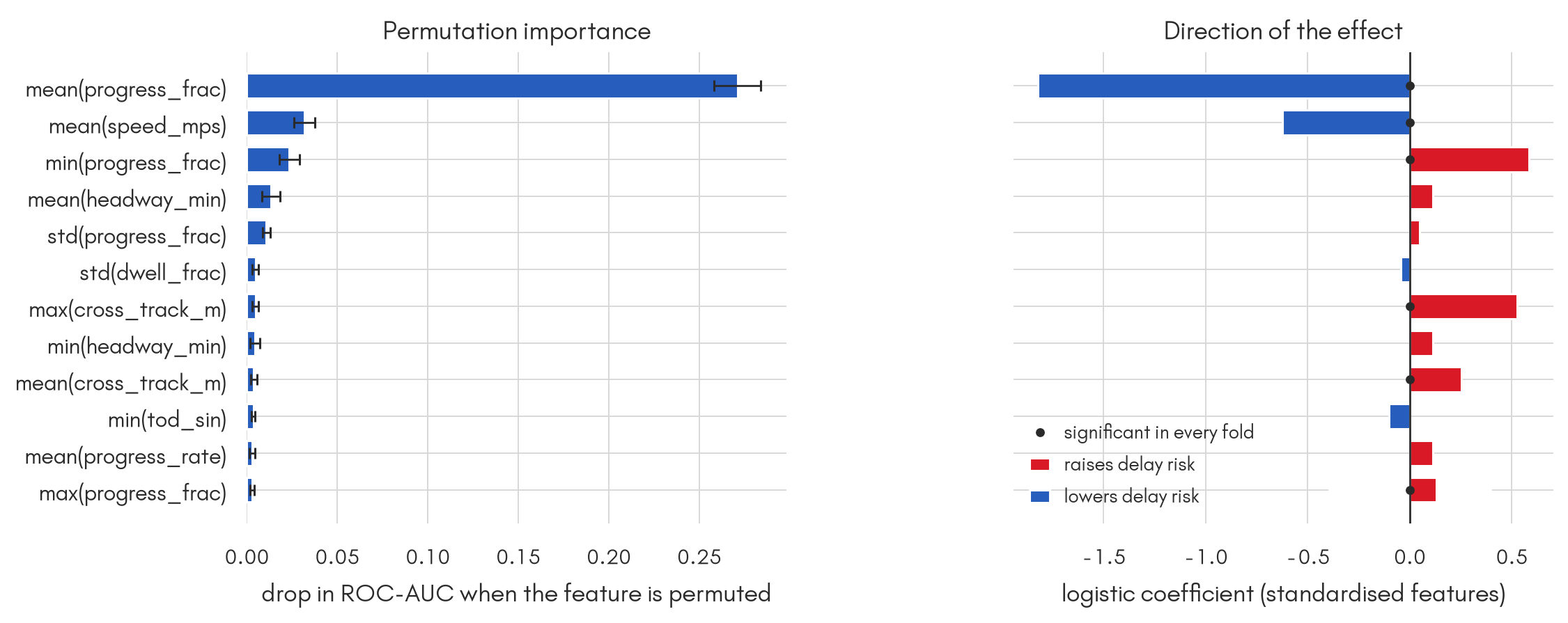}
\caption{What the delay model uses. Left, the drop in ROC-AUC when each feature
is permuted, averaged over the five folds with the cross-fold standard
deviation as the error bar. Right, the logistic coefficient on standardised
features, which gives the direction of each effect; a dot marks the features
that reach $p < 0.05$ in every fold.}
\label{fig:importance}
\end{figure*}

We take the best delay model,
RAW-STATS+XGBoost, whose 48 features are the mean, standard deviation, minimum
and maximum of each of the 12 input channels over the real timesteps, so every
feature has a plain physical reading. On the same day-grouped folds we compute
permutation importance~\cite{Fisher2019permutation} and SHAP
values~\cite{Lundberg2017shap}, and fit a logistic regression on the
standardised features to get coefficients with Wald standard errors and
$p$-values. Figure~\ref{fig:importance} shows the result.

Permuting \texttt{mean(progress\_frac)}, the average
along-route position of the bus across its time bins, costs $0.271$ ROC-AUC,
roughly eight times the next feature. Its coefficient is $-1.82$, so a trip
that sits further along its route on average is much less likely to be
labelled delayed. \texttt{mean(speed\_mps)} follows at $0.032$ with a
coefficient of $-0.62$, and \texttt{min(progress\_frac)} at $0.023$ with
$+0.58$. SHAP values give the same ordering for these three features and place
headway fourth, as permutation importance does.

The model is mostly detecting that a bus moved slowly and did not get far, which is close to what
being delayed means. It is a valid signal and it is learned without ever seeing
the schedule, but it is not a discovery about the causes of delay. The
contextual channels we expected to matter do not appear. Rainfall and the
time-of-day encodings are absent in the top ten except a single weak
\texttt{min(tod\_sin)} term, and headway, although fourth by permutation
importance, fails to reach significance in every fold. Only 10 of the 48
features are significant across all five folds.

\subsection{Interpretability of latent dimensions}

The same question can be put to the learned representation, so we correlated
each of the 64 CNN-AE latent dimensions with the delay label and fitted a
single-dimension logistic probe to each. No axis is individually meaningful,
since the strongest dimension reaches only $0.606$ ROC-AUC on its own and the
largest point-biserial correlation with the label is $|r| = 0.147$, with no
dimension above $0.20$, while all 64 together reach $0.771$. The representation spreads the
delay signal across many dimensions rather than isolating it on a few, so it
cannot be read axis by axis. This matches the delay-detection result, in that
the value of the latent space lies in compression and linear separability
rather than in interpretability.

\section{Service anomaly results}
\label{sec:anom}

\subsection{Agreement between representations}

\begin{figure}[t]
\centering
\includegraphics[width=0.86\columnwidth]{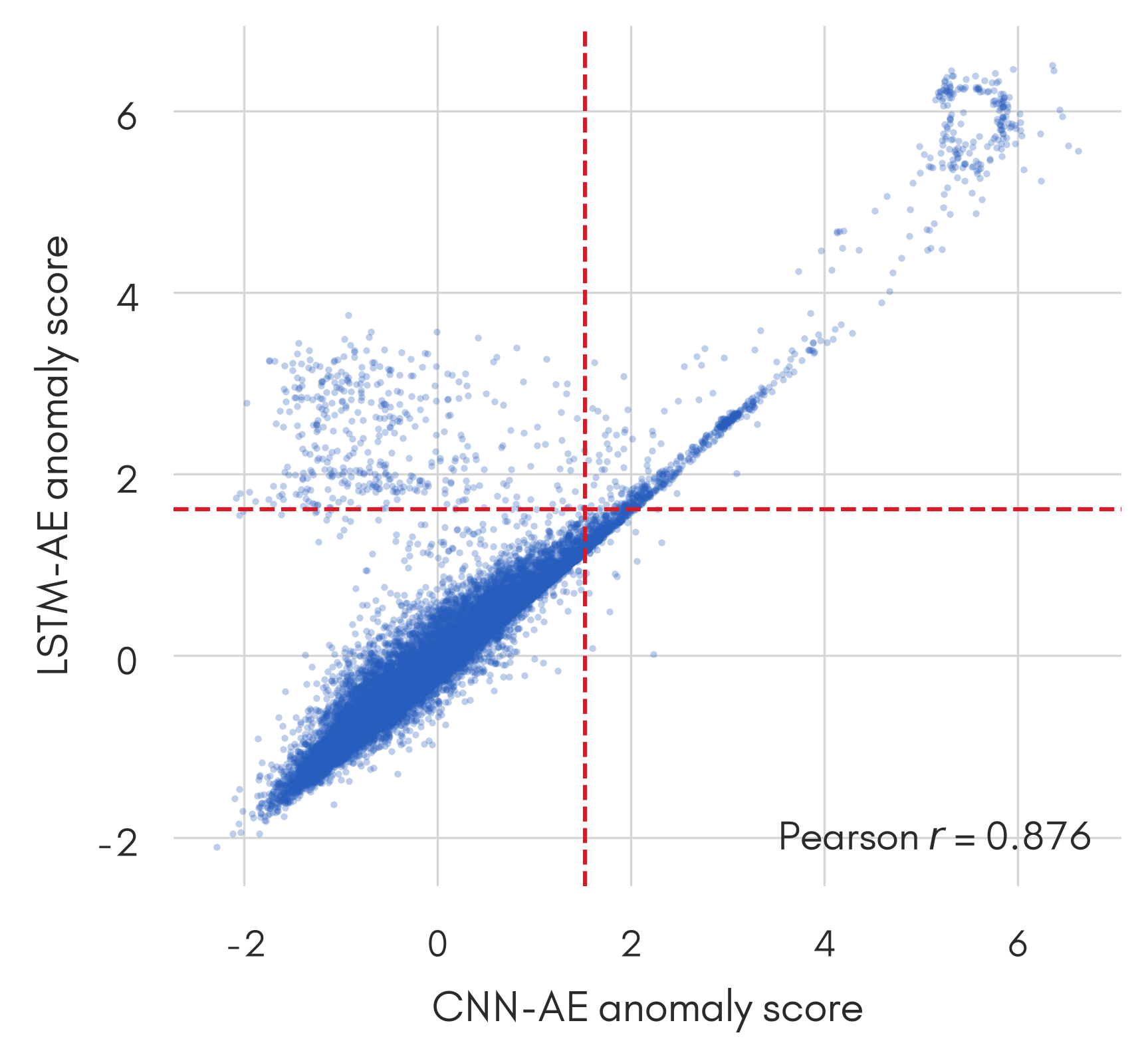}
\caption{CNN-AE versus LSTM-AE anomaly scores, standardised log reconstruction
error, so zero is the average trip. Dashed lines mark each model's 95th
percentile flagging threshold ($1.53$ for the CNN-AE, $1.61$ for the LSTM-AE).
Trips in the upper-right quadrant are flagged by both models, those in the
off-diagonal quadrants by only one.}
\label{fig:agree}
\end{figure}

The two AEs largely agree on what is abnormal. Their out-of-fold scores
correlate at Pearson $r = 0.876$ on the log scale (Spearman $\rho = 0.896$), and
of the $1{,}131$ trips each flags at the 95th percentile, $705$ are shared
(Jaccard $0.45$). Figure~\ref{fig:agree} shows the joint distribution.

\subsection{Circularity of the rainfall validation}
\label{sec:circular}

The anomaly score was initially validated against three external signals, namely
days with known degraded telemetry, rainfall, and an apparent citywide
disruption on 23 March. Rainfall, however, is \emph{itself an input channel} to
the autoencoder, so a rain-linked score may reflect reconstruction difficulty on
a rare input value rather than disrupted service.

We tested this directly by retraining the CNN-AE on the 11 non-weather channels
and repeating every validation. Table~\ref{tab:ablation} reports the result.

\begin{table*}[t!]
\centering
\caption{Weather-free ablation. Each validation is recomputed with the CNN-AE
retrained without the rainfall channel. Two of the three original signals do not
survive. The degraded-day row uses 20 and 22 March; see the text for 28 March.}
\label{tab:ablation}
\footnotesize
\tablefont
\begin{tabular}{lccc}
\toprule
Validation & With rainfall & Without rainfall & Verdict \\
\midrule
Degraded days (20, 22 Mar), median score
  & $+0.37$ vs $-0.21$ & $+0.50$ vs $-0.20$ & survives \\
\quad one-sided Mann--Whitney
  & $p = 9\times10^{-122}$ & $p = 6\times10^{-118}$ & \\
\midrule
Rain-hour flag rate vs dry
  & $12.2\%$ vs $4.4\%$ & $5.4\%$ vs $5.0\%$ & \textbf{does not survive} \\
\quad significance
  & $p = 2.8\times10^{-23}$ & $p = 0.84$ & \\
\midrule
23 Mar 15:30--17:00 window, median score
  & $+5.43$ & $-0.22$ & \textbf{does not survive} \\
\quad trips in global top-20
  & $20/20$ & $0/20$ & \\
\midrule
Day-level ordering (Spearman between variants)
  & \multicolumn{2}{c}{$\rho = 0.961$} & survives \\
\bottomrule
\end{tabular}
\end{table*}

\begin{figure}[b]
\centering
\includegraphics[width=0.935\columnwidth]{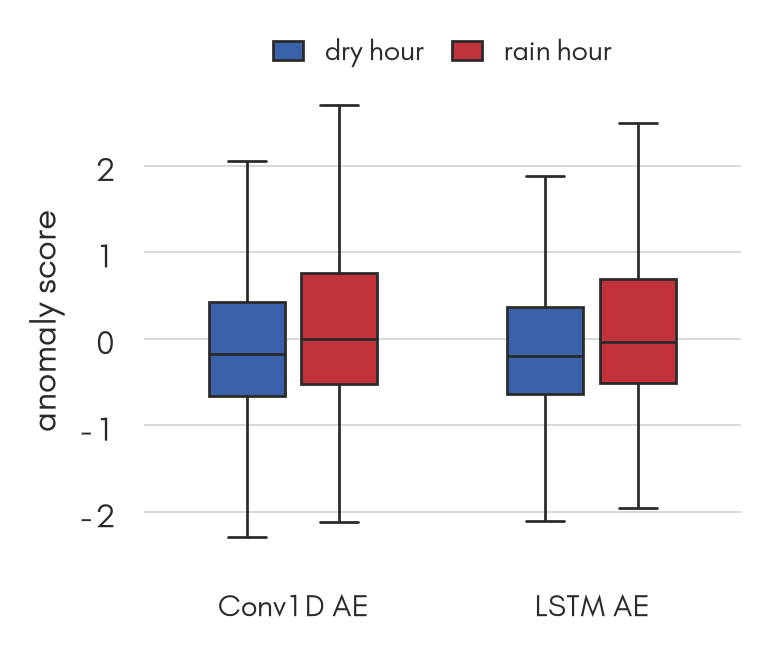}
\caption{Anomaly score by rain condition for the weather-inclusive models. The
association shown here is the one the ablation in Table~\ref{tab:ablation}
removes; it is reproduced to document what the original validation looked like.}
\label{fig:rain}
\end{figure}

The rain association vanishes once rainfall leaves the input. Rain-hour trips
are then flagged at $5.4\%$ against $5.0\%$ for dry hours ($p = 0.84$);
Figure~\ref{fig:rain} shows the association as it looked before the ablation,
under both weather-inclusive models. The
23 March ``citywide disruption'' vanishes just as completely. That window's
median score falls from $+5.43$ to $-0.22$, and none of its trips stay in the
global top 20. What we had read as a fleet-wide operational
event was the autoencoder failing to reconstruct a rare input value.

The third degraded day, 28 March, has only $449$ trips and scores \emph{below} the normal-day median
under both variants (weather-free day-median $-0.48$, rank 17 of 19), so
degraded telemetry does not always raise the reconstruction error. It is
excluded from that row and included in the weekend stratum of
Table~\ref{tab:strata}, where the effect is carried by 22 March.

A reconstruction error computed over a channel set that contains the checking
variable cannot establish an operational anomaly. The check and the input have
to stay separate.

\subsection{Effect of schedule-derived input channels}
\label{sec:schedchan}

The report keeps GTFS strictly on the label side. To show why, we added a
schedule-derived channel to the autoencoder input and retrained, in two
variants. The first uses \texttt{start\_deviation\_sec}, how early or late the
bus left, which an operator genuinely knows at dispatch and which is not the
label. The second uses \texttt{median\_delay\_sec}, the quantity the label is
thresholded from, as the strongest possible leak. Table~\ref{tab:schedchan}
reports both.

\begin{table}[t]
\centering
\caption{Adding a schedule channel to the AE input. Delay detection becomes
close to trivial, while the reconstruction error stays uninformative about the
label. Mean over the five folds.}
\label{tab:schedchan}
\footnotesize
\setlength{\tabcolsep}{4pt}
\tablefont
\begin{tabular}{lccc}
\toprule
AE input & ROC-AUC & PR-AUC & recon.\ AUC \\
\midrule
12 ch., no schedule & $0.838$ & $0.629$ & $0.505$ \\
$+$ start deviation      & $0.961$ & $0.887$ & $0.519$ \\
$+$ median stop delay    & $0.989$ & $0.964$ & $0.515$ \\
\bottomrule
\end{tabular}
\end{table}

Both variants inflate delay detection enormously, from $0.838$ to $0.961$ and
then to $0.989$ ROC-AUC. The second number is the expected one, since a model handed
the quantity its label is computed from recovers that label almost perfectly,
which measures arithmetic rather than representation quality. The first is more
interesting, because start deviation is legitimate operational data, and it
still lifts the score by $0.12$. A bus that leaves late tends to stay late, so
including it turns the task into something much closer to reading the schedule
than to learning from telemetry.

We expected the reconstruction error to partly reproduce the delay label once schedule
information entered the input. It does not, and the error's ability to rank delayed
trips stays at chance in all three settings ($0.505$, $0.519$, $0.515$). The
damage falls on the supervised side, where the classifiers read the schedule
straight out of the latent vector, rather than on the unsupervised score. The two tasks
do stop being independent, but by a different route than we assumed.

\subsection{A single global autoencoder}
\label{sec:globalae}

Training a separate AE inside each fold gives five encoders, five error scales
that must be standardised before pooling, and the five artificial lobes in
Figure~\ref{fig:tsne}. The alternative is a single AE fitted once on a
historical slice, with cross-validation only on the classifiers. We
trained the AE on the earliest $40\%$ of service dates and cross-validated the
classifiers on the later days, keeping the two sets of days disjoint so nothing
the AE saw can reach a classifier test fold.

GLOBAL-AE+XGBoost reaches $0.833 \pm 0.021$
ROC-AUC and $0.616 \pm 0.029$ PR-AUC, against $0.838 \pm 0.016$ and
$0.629 \pm 0.051$ for the per-fold arrangement, a difference well inside the
cross-fold spread and obtained while training on only $40\%$ of the days. What
it buys is structural, since the reconstruction error now lives on a single scale
(history mean $0.3150$, held-out mean $0.4867$), so the within-fold
standardisation step of Section~\ref{sec:method}.\ref{sec:anomscore} is
no longer needed and the fold lobes of Figure~\ref{fig:tsne} cannot arise. For a deployed
system this is the better arrangement, since it matches how the model would
be used, fitting the representation on history and applying it going
forward.

\subsection{Day type versus service disruption}
\label{sec:daytype}

\begin{figure*}[t!]
\centering
\includegraphics[width=\textwidth]{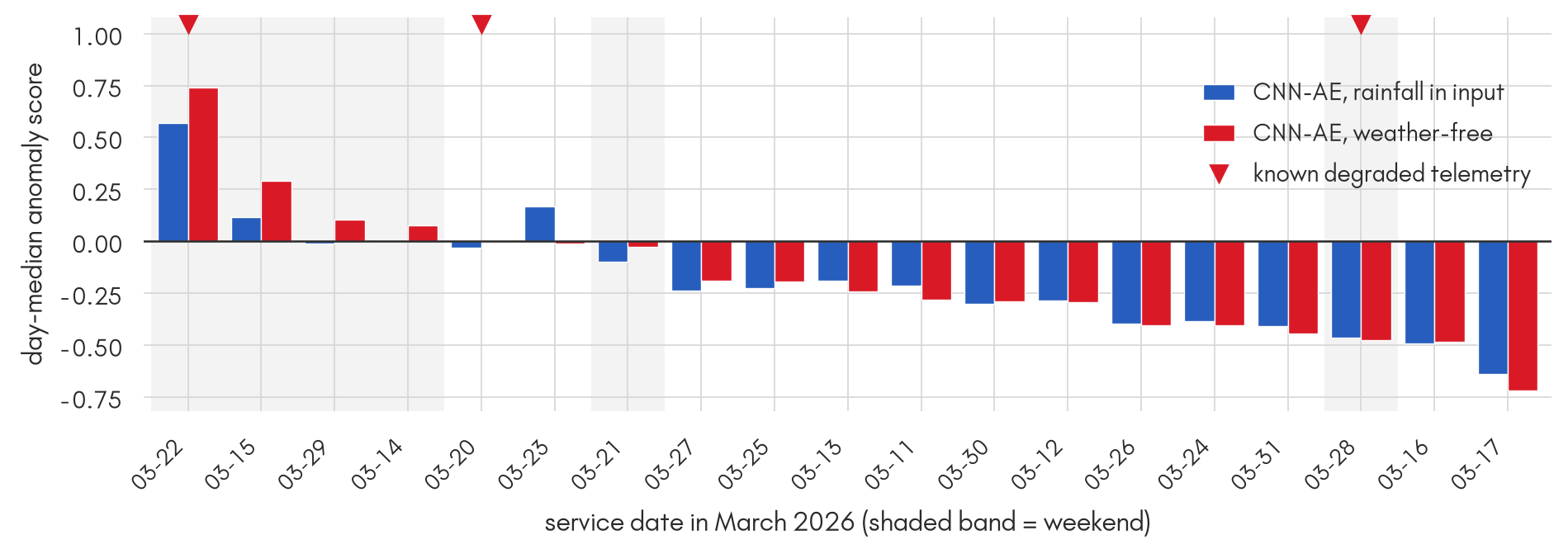}
\caption{Day-median anomaly score, sorted by the weather-free variant. Weekends
(shaded) occupy five of the top seven positions.}
\label{fig:daylevel}
\end{figure*}

Having removed the weather confound, we examined what drives the
day-level ordering. Figure~\ref{fig:daylevel} shows day-median scores for both
variants with weekends shaded and known degraded days marked.

The dominant factor is \textbf{day type} rather than disruption. Weekend days occupy
ranks 1, 2, 3, 4 and 7 of 19. At trip level the weekend median score is $+0.178$
against $-0.286$ for weekdays (one-sided Mann--Whitney
$p = 1.2\times10^{-205}$), with flag rates of $6.1\%$ and $4.6\%$. This is
expected, since Sunday service genuinely differs in frequency,
routing and dwell behaviour. It does mean that a plain reading of the day
ranking would mistake ``Sunday'' for ``disrupted''.

The degraded-telemetry signal nevertheless survives stratification.
Table~\ref{tab:strata} separates the two effects.

\begin{table*}[t!]
\centering
\caption{Degraded versus normal days, stratified by day type (weather-free
CNN-AE). The degraded-day effect holds within both strata, so it is not an
artefact of 22 March being a Sunday.}
\label{tab:strata}
\small
\tablefont
\begin{tabular}{lccc}
\toprule
Stratum & Degraded median & Normal median & one-sided $p$ \\
\midrule
Weekdays & $+0.002$ ($n=612$)    & $-0.295$ ($n=15{,}860$) & $9.5\times10^{-8}$ \\
Weekends & $+0.418$ ($n=1{,}494$) & $+0.110$ ($n=4{,}639$)  & $4.1\times10^{-14}$ \\
\midrule
22 Mar vs other Sundays & $+0.740$ ($n=1{,}045$) & $+0.206$ ($n=2{,}005$) & $5.2\times10^{-42}$ \\
\bottomrule
\end{tabular}
\end{table*}

22 March remains the highest-scoring day even when compared only against other
Sundays ($+0.740$ vs $+0.206$, $p = 5.2\times10^{-42}$). Its telemetry volume
was $65{,}993$ GPS rows against about a million on a normal weekday and half a
million on the other two Sundays, by far the most severe drop in the period.

\subsection{What the score detects}

Combining the ablation and the stratification, we can now separate the three
phenomena that the reconstruction error had been mixing together.

\Needspace*{13\baselineskip}
\begin{enumerate}
\item \textbf{Telemetry degradation.} Detected, and separable from day type.
      This is a \emph{data-quality} problem rather than a service problem, so it
      works as a positive control that shows the score is sensitive. It says
      nothing about whether the method finds operational faults.
\item \textbf{Rare input values (weather).} Detected, but only in a circular
      way. The signal disappears once the variable leaves the input, so it
      carries no operational meaning.
\item \textbf{Genuine operational disruption.} \textbf{Not demonstrated.} We have
      no validated instance of the score detecting a real service disruption.
      The one candidate (23 March) was an artefact.
\end{enumerate}

The anomaly score is, on this evidence, a detector of \emph{unusual days}. What makes
a day unusual is mostly its day type, and secondarily any telemetry
degradation. Establishing that it detects service disruption requires
ground-truth disruption labels, which this dataset does not carry.

\subsection{Threshold calibration}
\label{sec:threshold}

The choice of a global 95th-percentile threshold merits explanation.
The percentile sets an \emph{inspection budget} rather than estimating an
anomaly rate. At $5\%$ of $22{,}605$ trips it gives roughly 57 alerts per service
day, a number an operator can review. There is no ground truth against which
$5\%$ could be calibrated as a rate.

We compared four rules on the existing weather-free scores. No retraining is
required; this is pure post-hoc thresholding.

The first, \emph{global-p95}, is the current rule, one pooled 95th percentile.
The second, \emph{dynamic}, is an absolute cut at $\mu + 1.645\sigma$
calibrated leave-one-day-out. The third and fourth
re-standardise the score before applying that same cut, \emph{per-line} within
each bus line and \emph{per-hour} within each hour of the day.
Table~\ref{tab:thresh} compares the four.

\begin{table}[htbp]
\centering
\caption{Threshold rules evaluated on the weather-free CNN-AE scores. The
diagnostic case is 22 March, the most severely degraded day.}
\label{tab:thresh}
\footnotesize
\setlength{\tabcolsep}{4pt}
\tablefont
\begin{tabular}{lccc}
\toprule
Rule & Overall & 22 Mar & Lift \\
\midrule
global-p95 & $5.0\%$ & $5.7\%$  & $1.1\times$ \\
dynamic    & $6.8\%$ & $10.3\%$ & $1.5\times$ \\
\textbf{per-line}   & $6.1\%$ & $\mathbf{29.3\%}$ & $\mathbf{4.8\times}$ \\
per-hour   & $6.6\%$ & $10.3\%$ & $1.6\times$ \\
\bottomrule
\end{tabular}
\end{table}

The global percentile is \emph{dominated by differences between bus lines}. Some lines simply reconstruct
worse than others, so a pooled cut spends most of its alarm budget on those
lines whatever the day. On the most severely degraded day in the dataset it
flags $5.7\%$ of trips, essentially the baseline rate.

\begin{figure*}[t]
\centering
\includegraphics[width=\textwidth]{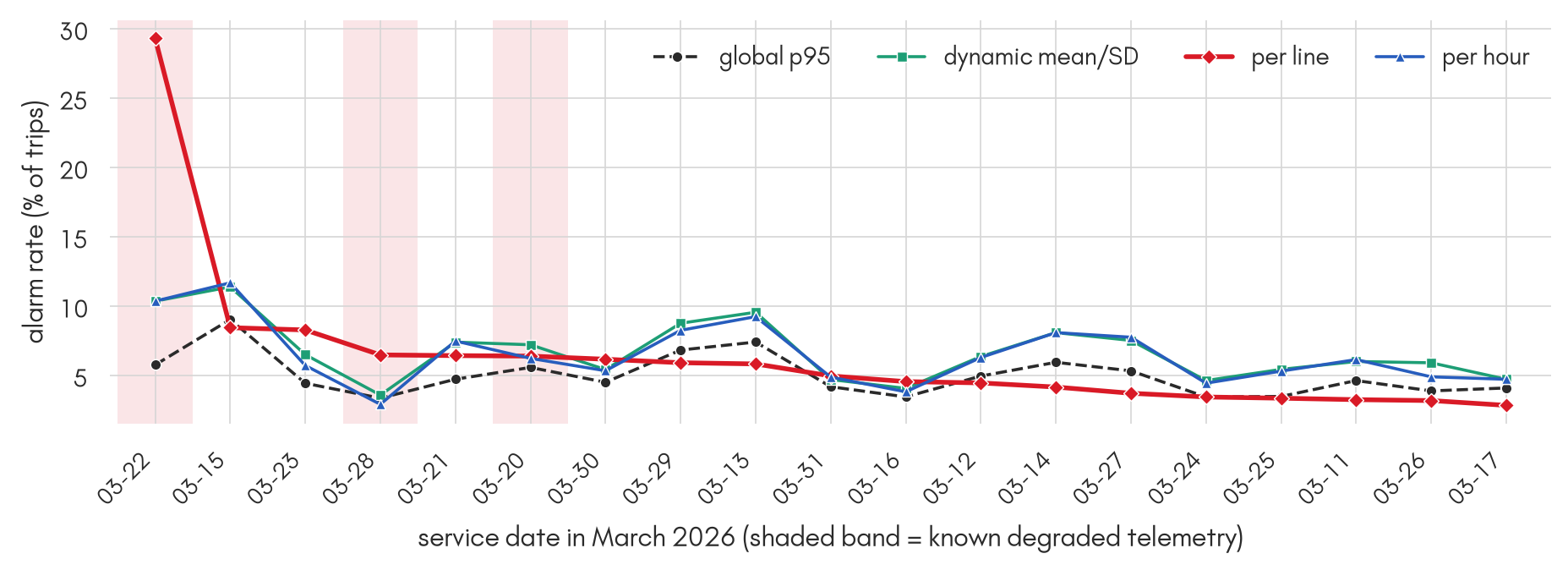}
\caption{Alarm rate per service date under the four threshold rules. Known
degraded days shaded. Per-line calibration is the only rule that makes 22 March
stand out.}
\label{fig:thresh}
\end{figure*}

Re-standardising each score within its own line raises that to $29.3\%$, a
$4.8\times$ lift over its own overall rate of $6.1\%$. The dynamic mean/SD
threshold helps far less ($1.5\times$) than per-line calibration does. Figure~\ref{fig:thresh}
shows the full per-day picture.

In practice, then, the alarm threshold limits detection far more than the
score does. An operator running this system would gain more from per-line
scaling than from any change of representation we tested.

\section{Network-level distribution of delay}
\label{sec:e10}

The analyses so far treat trips as interchangeable. This section asks where in
the network the delays sit, whether passenger demand explains them, and whether
delay spreads from one trip to the next.

\begin{figure*}[t]
\centering
\includegraphics[width=\textwidth]{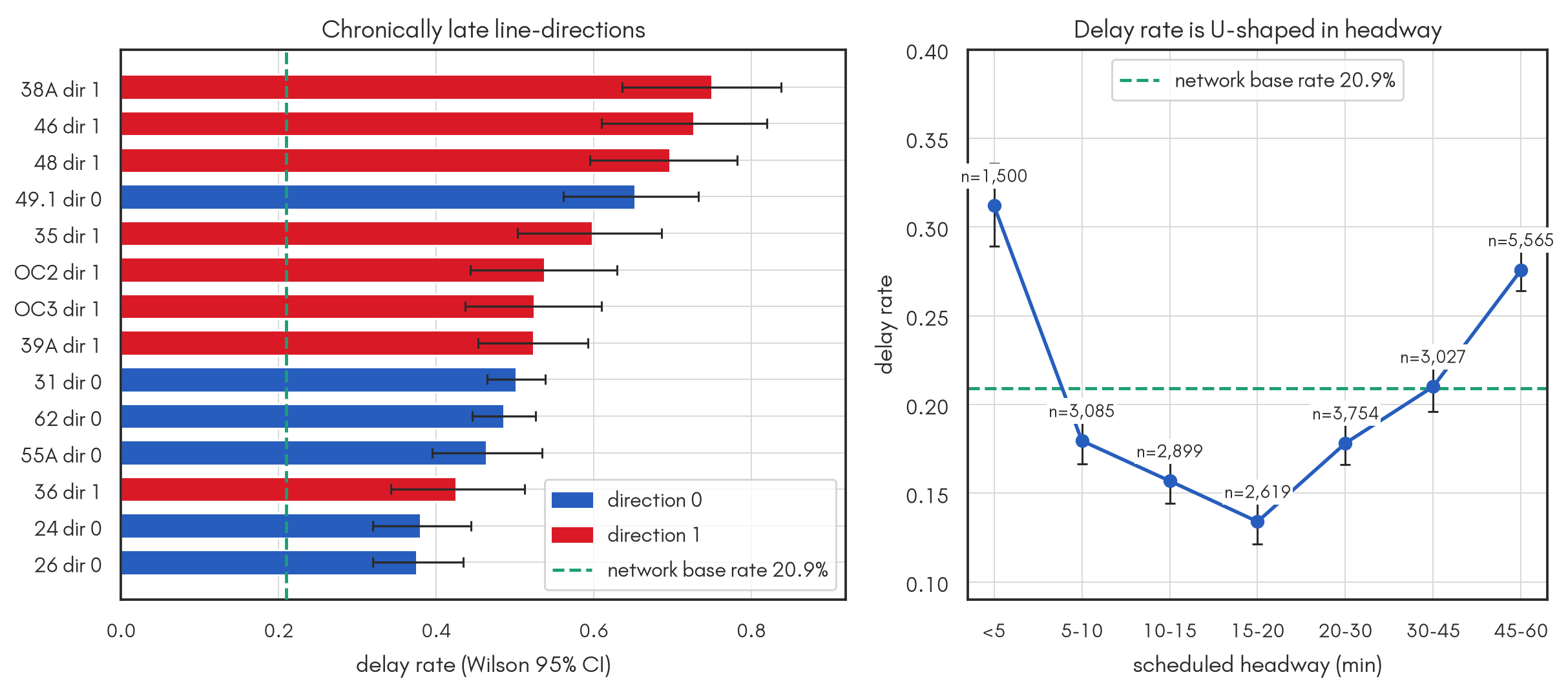}
\caption{Where the delay sits in the network. Left, the 14 worst of the 22
chronically late line-directions, with delay rate and its Wilson $95\%$
interval, coloured by direction of travel; the dashed line is the network rate
of $20.9\%$. The two directions of a line are ranked separately and often fall
far apart. Right, delay rate against scheduled headway, with the number of
trips in each band. The relation is U-shaped, so a single rank correlation
describes it poorly.}
\label{fig:chronic}
\end{figure*}

\subsection{Concentration by line-direction}

Against a network delay rate of $20.9\%$, individual lines range from $1.4\%$
to $48.8\%$, a factor of $35$. Thirteen of the $41$ lines with at least $50$
trips are chronically late in the sense that the Wilson~\cite{Wilson1927} lower
bound of their delay rate sits above the network rate; the worst are line 55A at $48.8\%$
($95\%$ CI $42.3$--$55.5$), line 31 at $46.1\%$ and line 62 at $44.0\%$.

Aggregating by line hides the real structure, because the two directions of the
same line often behave nothing alike. Line 48 is delayed $13.6\%$ of the time in
one direction and $69.7\%$ in the other; line 35 runs $8.2\%$ against $59.8\%$.
The worst line-directions, 38A direction 1 at $75.0\%$ and 46 direction 1 at
$72.7\%$, carry median delays of eight to twelve minutes, so these are not
marginal cases sitting just over the five-minute boundary.
Figure~\ref{fig:chronic} (left) ranks them with their intervals.

Concentration follows the same pattern, since ranked by line the top ten carry
$51.4\%$ of delayed trips but also $45.5\%$ of all trips, which is barely any
concentration at all. Ranked by line-direction, $22$ chronic pairs account for
$27.3\%$ of trips and $50.7\%$ of every delayed trip in the dataset, an
enrichment of $1.86\times$, with an internal delay rate of $38.9\%$. Half the
delay problem lives in about a quarter of the service, and it is addressable by
direction.

\subsection{Passenger demand}

We aggregated all $6{,}790{,}681$ boarding transactions into $30{,}136$
line-by-hour cells and attached them to $85.7\%$ of trips. A per-trip join is
impossible: the ticketing \texttt{vehicle\_number} has no overlap with the AVL
vehicle identifier, and the ticket records carry no direction field.

Taken at face value the demand features help a little, lifting PR-AUC from
$0.6370 \pm 0.0528$ to $0.6467 \pm 0.0537$, an improvement in all five folds
($p = 0.013$), with ROC-AUC moving $+0.0029$ ($p = 0.110$). The control changes
the reading completely. A single integer identifying the line beats the whole
demand block ($+0.0124$ PR-AUC, $p = 1.0 \times 10^{-4}$), and once that line
identifier is present the demand features add nothing ($-0.0012$ ROC-AUC,
$p = 0.465$; $-0.0018$ PR-AUC, $p = 0.568$). The apparent gain was demand
acting as a proxy for which line the trip belonged to. At the aggregation the
join permits, $567$~MB of ticketing data adds no predictive power beyond a line
label.

\subsection{Delay propagation}

Consecutive trips on the same line and direction look strongly coupled, with
$50.6\%$ of trips following a delayed predecessor being delayed themselves
against $13.1\%$ otherwise, a relative risk of $3.86$. Almost all of that is
shared conditions rather than transmission. The only mechanism this data can
identify is a hand-over between consecutive trips run by the same
vehicle, and there the excess disappears, since a within-line-day permutation test
gives $z = -0.77$ ($p = 0.826$) and a Mantel--Haenszel~\cite{Mantel1959} odds
ratio of $0.829$.
The residual association is carried by pairs run by \emph{different} vehicles
and vanishes once the hour of day enters the stratum, which is what a common
cause looks like rather than a cascade.

Headway behaves non-monotonically, with $31.2\%$ of trips delayed below five
minutes of headway, $13.4\%$ between fifteen and twenty, and $27.6\%$ again
between forty-five and sixty, the U-shape in Figure~\ref{fig:chronic} (right).
A Spearman correlation of $+0.060$ is significant and close to meaningless on
such a relationship. Lines that share stops do
co-vary more than lines that do not (mean correlation $+0.139$ against $+0.069$,
Mann--Whitney $p = 0.0010$), so spatial structure exists. Co-variation across a
corridor is not by itself evidence of transmission along it, but the timing
distinguishes the two. Aggregating each line to hourly delay rates over $3{,}177$
line-hour cells and correlating every ordered pair of the $23$ lines with enough
hours, the same-hour correlation is $+0.163$ while the same pairs at one hour of
lag fall to $+0.126$, a drop of $0.037$ (paired $t$ over $451$ pairs,
$p = 1.6\times10^{-5}$). The smallest drop this design could detect at $80\%$
power is about $0.024$, so the comparison is adequately powered and the effect
is about one and a half times that floor. Lines move
together in the same hour rather than one leading another, which is the
signature of a shared cause such as traffic or weather rather than delay
travelling along a corridor. Together with the null result on same-vehicle
hand-over, the evidence points consistently at common conditions rather than
propagation.

\subsection{Operational implications}
\label{sec:act}

Delay is concentrated by line-direction rather than spread evenly, the anomaly
threshold works far better when calibrated per line
(Section~\ref{sec:anom}.\ref{sec:threshold}), and most of the delay signal is
present after a quarter of the route
(Section~\ref{sec:delay}.\ref{sec:early}). Together these three findings
describe a per-line
early-warning monitor rather than a network-wide one, a monitor that scores
trips against their own line's normal behaviour, acts before the trip completes,
and concentrates attention on the twenty-odd line-directions that carry half
the problem. What
the evidence does not support is a demand-driven intervention, since demand adds
nothing once the line is known, or a cascade-control strategy, since we could
not detect transmission between consecutive trips.

\section{Discussion}
\label{sec:disc}

\subsection{Supported claims}

One label-free representation can indeed be learned once and reused for both a
supervised and an unsupervised task, and that reuse works. For delay detection,
however, the representation is not what moves performance. Raw per-trip
statistics with gradient boosting are at least as accurate, and the
representation earns its place only through linear separability, which matters
when a linear model is required. The day-level anomaly signal is the stronger
result. It survives the choice of autoencoder, appearing under three separately
trained encoders, one of which sees no weather at all, and it is partly shared
by a contrastive encoder with no reconstruction objective. Per-line threshold calibration is worth
more than any change of representation we measured.

Outside that framing, the delay model turns out to rest almost entirely on one
feature, the average along-route position of the bus,
which carries roughly eight times the importance of the next and amounts to
detecting that a bus moved slowly and did not get far. That is a valid signal
learned without ever seeing the schedule, but it is close to a restatement of
what being delayed means rather than an explanation of it, and the contextual
channels we expected to matter, rainfall and time of day, do not appear. The
learned representation is no more interpretable, since no single latent
dimension exceeds $0.61$ ROC-AUC on its own while all 64 together reach
$0.771$, so the signal is distributed rather than axis-aligned. At the network level, delay
concentrates by line-direction rather than by line, with 22 chronic pairs
carrying half of all delayed trips from a quarter of the service, while
$6.8$ million boarding transactions add nothing once the line identity is known
and trip-to-trip propagation cannot be detected at all.

Three claims that might be read into this work are not supported by it. We have
no validated case of the method detecting a genuine service disruption. The
anomaly score is substantially a day-type signal, so it should not be read as a
mainly operational one. And the latent representation does not improve delay
prediction accuracy.

\subsection{Deployment conditions}

The pipeline is worth its extra
complexity when (i) schedule data is unreliable or
absent, so the deterministic comparison is unavailable; (ii) a single
representation must serve several downstream tasks, so one encoder serves all of them; or (iii) a linear downstream model is required. Where a reliable
GTFS feed exists and only retrospective delay classification is needed, the
deterministic comparison is simpler and the raw-feature baseline is more
accurate.

\subsection{Practical use}
\label{sec:practical}

Three parts of this work transfer directly to an agency running the same kind of
feed. The first is the labelling pipeline. A raw GPS archive carries no delay
annotations, and Section~\ref{sec:prep} derives them from a dated GTFS snapshot
by map-matching each trace to its route shape and interpolating a stop arrival
wherever along-route progress crosses a stop, which turns an unlabelled archive
into a supervised dataset with no manual annotation. The gate thresholds in
Table~\ref{tab:gates} and the coverage rule of
Section~\ref{sec:prep}.\ref{sec:gatecheck} are the part worth copying, since
they decide how much of the archive survives and we report what each one costs.

The second is the monitor of Section~\ref{sec:e10}.\ref{sec:act}, whose cheapest
half is the threshold rule. Re-scaling each trip's anomaly score within its own
bus line, instead of against one network-wide cut, lifts the alarm rate on the
worst degraded day from $5.7\%$ to $29.3\%$. That is a few lines of arithmetic
over scores an agency already has, and it beat every change of representation we
tried. Scoring a trip from its first quarter keeps $93.8\%$ of the
discriminative power of the finished trip, so the same monitor can raise an
alert while the bus is still running, which a schedule comparison cannot do
because the median stop delay it needs is undefined until the trip ends.

The third is a target list of $22$ line-directions, which carry half of all
delayed trips from just over a quarter of the service. The two directions of one
line often behave nothing alike, with line 48 delayed $13.6\%$ of the time one
way and $69.7\%$ the other, so scheduling and field attention is better spent by
direction than by line.

The negative results are worth as much, because each one closes off an expensive
avenue. An agency does not need to join ticketing data to predict delay, since
$6.8$ million boarding transactions add nothing once the line is known. It
should not build a cascade-control strategy on this evidence, since we could not
detect delay transmission between consecutive trips. It does not need
representation learning to classify delay, since 48 summary statistics with
gradient boosting match the autoencoder. And the circularity we ran into
generalises well past transit, in that an anomaly score cannot be validated
against a variable that is also one of its own inputs, because the check then
confirms nothing beyond the model's difficulty with rare values.

\subsection{Limitations}
\label{sec:limits}

The central limitation is that we have no ground truth for disruption, so every
anomaly check here is a proxy. The day-level conclusions rest on 19 to 20
service dates, six of them weekends, and the paired tests run at $n = 5$, where
the Wilcoxon floor of $p = 0.031$ stops us resolving anything finer than
``holds in every fold''. All of it comes from one city in one month, March 2026
in Niter\'oi, which happens to include an unusual stretch of degraded
telemetry. We tested whether the quality gates bias the anomaly analysis and found no
support for it (Section~\ref{sec:prep}.\ref{sec:gates}). The remaining gates
were relaxed collectively rather than individually, so their separate
contributions are not isolated. A
nested search (Section~\ref{sec:delay}.\ref{sec:tuning}) moves the metrics
slightly and leaves the ordering between raw and latent features intact, and a sweep over
the autoencoder's own latent dimension and depth
(Section~\ref{sec:delay}.\ref{sec:robust}) does not change it
either.

\section{Conclusion}

We aimed to show that one unsupervised representation could serve both delay detection and service anomaly detection. The representation does serve both, but the evidence does not support the stronger claim we began with.

For delay detection, gradient boosting on a per-trip summary of the same twelve channels matches, or slightly outperforms, the learned representation ($0.846$ vs $0.838$ ROC-AUC). This difference is not statistically meaningful on the strict dataset, but becomes significant when the preprocessing filters are relaxed. The latent space earns its place only through linear separability.

For anomaly detection, two of our three original checks turned out to be
circular, because rainfall served as an input channel and as the external check
at the same time. Removing it from the input erased
both the rain association and an apparent citywide disruption. What remains is a
day-level signal that is stable across three independently trained
autoencoders but is driven mostly by day type (weekends hold five of the
top seven positions), with telemetry degradation separable within strata. We
have no validated instance of the score detecting genuine service disruption.

The delay model is dominated by a single feature amounting to ``the bus moved
slowly and did not get far'', and no latent dimension is individually
interpretable, so neither the raw nor the learned representation explains delay
so much as detects it. At the network
level the problem is concentrated: 22 chronic line-directions carry half of all
delayed trips, while passenger demand adds nothing beyond the line label and
trip-to-trip propagation is undetectable.

The most useful result concerns the alarm threshold, where re-scaling scores
within each bus line lifts the alarm rate on the worst degraded day from
$5.7\%$ to $29.3\%$, a $4.8\times$ gain over its own baseline rate that no change of
representation we tested came close to matching. Taken with the concentration by line-direction and the
availability of most of the signal after a quarter of the route, this points at
a per-line, early-warning monitor as the deployable form of the work.

\section*{Data and code availability}
The dataset was obtained under the NetMob 2026 Data Challenge terms and is
available from the organisers to admitted participants. Weather data are from
the public municipal feed.

All analysis code is openly available at
\url{https://github.com/mttgermano/Netmob2026}. The dataset repository is at
\url{https://github.com/lprm-ufes/Netmob2026}.

The numbers in this report can be reproduced from the artifacts in
\texttt{code/artifacts\_*} and the \texttt{exp\_*.py} scripts. The specific scripts behind the results here are
\texttt{ablation\_norain.py} (Section~\ref{sec:anom}.\ref{sec:circular}),
\texttt{threshold\_calib.py} (Section~\ref{sec:anom}.\ref{sec:threshold}),
\texttt{baseline\_raw.py} (Section~\ref{sec:delay}) and
\texttt{make\_report\_figs.py} (Figures~\ref{fig:label}, \ref{fig:recon},
\ref{fig:importance}, \ref{fig:agree}, \ref{fig:rain}, \ref{fig:daylevel}
and \ref{fig:thresh}), \texttt{exp\_e10\_fig.py}
(Figure~\ref{fig:chronic}) and \texttt{03\_anomaly\_analysis.ipynb}
(Figure~\ref{fig:tsne}). \texttt{exp\_gatebias\_leadlag.py} covers the
gate-bias checks of Section~\ref{sec:prep}.\ref{sec:gates} and the lead--lag
comparison of Section~\ref{sec:e10}.

\bibliography{refs}

\end{document}